\documentclass[journal]{IEEEtran}
\usepackage{graphicx}
\usepackage{amssymb}
\usepackage{cite}
\usepackage{soul}
\usepackage{color}

\usepackage{xcolor}
\usepackage{amsthm}
\usepackage{framed}
\colorlet{shadecolor}{yellow!50}

\ifCLASSINFOpdf

\fi

\begin{document}

\title{Adaptive Neuro-Fuzzy Control of a Spherical Rolling Robot Using Sliding Mode Control Theory-Based Online Learning Algorithm}

\author{Erkan~Kayacan~\IEEEmembership{Student Member}~,~Erdal~Kayacan~\IEEEmembership{Senior Member}~,~Herman~Ramon~\IEEEmembership{}~and~Wouter~Saeys~\IEEEmembership{}% <-this % stops a space
\thanks{E. Kayacan, E. Kayacan, H. Ramon ans W. Saeys are with the Division of Mechatronics, Biostatistics and Sensors, Department of Biosystems, KU Leuven, Kasteelpark Arenberg 30, B-3001 Leuven, Belgium.
e-mail: \{erkan.kayacan, erdal.kayacan, herman.ramon, wouter.saeys\}@biw.kuleuven.be }
}

\markboth{\textbf{PREPRINT VERSION: }IEEE TRANSACTIONS ON CYBERNETICS, vol.43, no.1, pp.170-179, Feb. 2013.}
{Shell \MakeLowercase{\textit{et al.}}: Bare Demo of IEEEtran.cls for Journals}
\maketitle

\begin{abstract}
As a model is only an abstraction of the real system, unmodeled dynamics, parameter variations and disturbances can result in poor performance of a conventional controller based on this model. In such cases, a conventional controller cannot remain well-tuned. This paper presents the control of a spherical rolling robot by using an adaptive neuro-fuzzy controller in combination with a SMC theory-based learning algorithm. The proposed control structure consists of a neuro-fuzzy network and a conventional controller which is used to guarantee the asymptotic stability of the system in a compact space. The parameter updating rules of the fuzzy-neuro system using SMC theory are derived, and the stability of the learning is proven using a Lyapunov function. The simulation results show that the control scheme with the proposed SMC theory-based learning algorithm is able to not only eliminate the steady state error but also to improve the transient response performance of the spherical rolling robot without knowing its dynamic equations.
\end{abstract}

%\begin{IEEEkeywords}
%Adaptive neuro-fuzzy control, Sliding mode learning algorithm, Spherical rolling robot.
%\end{IEEEkeywords}

\IEEEpeerreviewmaketitle

\section{Introduction}

\IEEEPARstart{M}{obil robots} are increasingly used in a variety of non-industrial applications such as security surveillance, search and rescue, children education, entertainment, etc. Spherical rolling mechanisms exhibit a number of advantages with respect to wheeled and legged mechanisms. All mechanical and electrical components including the actuation mechanism are securely located inside a spherical shell rolling itself over the ground surface. The motion of a sphere rolling without slipping over a surface is governed by non-holonomic constraints, therefore spherical rolling robots are classified as non-holonomic mechanical systems.

The fundamental difference between spherical and wheeled rolling motions is the instantaneous number of degrees of freedom (DOF) between the mobile body and ground surface. Since the sphere can simultaneously rotate around the transverse and longitudinal axes, its instantaneous mobility is greater than that of a wheel. Besides, spherical rolling mechanisms can change their direction of motion easier than wheeled mechanisms. Unlike wheeled and legged mechanisms, spheres can not fall over. The general problem of stability of equilibrium frequently encountered in mobile robotics is naturally avoided with the use of spherical rolling. On the other hand, highly complex nonlinear equations are needed to describe the dynamics of spherical rolling systems.

A number of spherical robot prototypes have been developed in recent years. One of the present mechanical structures consists of a wheeled vehicle located at the bottom of the sphere. In such a scheme, the motion is provided by the interaction of the vehicle's wheels and the sphere. The wheeled vehicle's motion inside the sphere being also non-holonomic, the overall system is the combination of two non-holonomic mechanical systems \cite{Dias}. A linear model for the longitudinal dynamics of the vehicle has been proposed to simplify the equations of motion of the system \cite{Camicia}. Since spherical rolling robots have highly nonlinear and very complex dynamics, the linearization approach is not proper for the study of these systems. A similar design utilizes a single wheel resting on the bottom of the sphere \cite{Halme}.

In another mechanical design, the driving motion of the rolling robot is obtained by changing the position of the mass center inside the sphere, which provides a gravitational torque \cite{Javadiasme,Mukherjee2}. Such as design has been proposed in \cite{Joshi,Joshi2} where two DC motors associated with two flywheels are mounted inside the sphere. When one of the flywheels turns around an arbitrarily chosen reference direction, the spherical rolling robot rolls in the opposite direction due to the conservation of angular momentum. However, such a mechanism cannot roll over curvilinear trajectories. In order to change its direction of motion, the sphere must stop rolling and then turn left or right around the vertical axis. A similar design utilizes two perpendicular rotors mounted inside the sphere \cite{Bhattacharya}.

A spherical robot with one DOF pendulum as actuation mechanism was designed in \cite{Ming}. A simplified dynamic model of a spherical robot with one DOF pendulum is derived under some assumptions in \cite{Jia,Liu}. One of these assumptions is that the longitudinal and lateral motions are decoupled from each other. Motion control has been established by feedback linearization considering the decoupled dynamics \cite{Liu2} and only driving motion \cite{Liu3}.

Having different mechanical designs, spherical robots are used for very different purposes. For example, Michaud \cite{Michaud,Michaud2} invented a spherical mobile robot called Roball, which is a tool in child-development studies. Roball, an autonomous robot, contributes to the development of children's language, affective, intellectual and social skills. Longitudinal and lateral models of Roball are discussed in \cite{Laplante}.

Due to the highly nonlinear equations constituting the dynamic, a kinematic model of the spherical rolling robots is usually used for path planning and in the design of controllers. A trajectory was simulated using a kinematic model in \cite{Alizadeh}. Under some assumptions, a feedback controller for a kinematic model and a dynamic model based on the backstepping method are also designed \cite{Otani}.

In most of the papers in literature, some assumptions are made to obtain simplified dynamic models for spherical rolling robots. For example, feedback control and feedback linearization have been applied to the system by using these simplified dynamic models \cite{Erkan}. Furthermore, the linearized dynamic model is also obtained and some linear control methods are applied to the system. However, these approaches are not feasible in practice due to the highly complex nonlinear equations. Moreover, although the kinematic model has been used to control the system because of its simplicity, it does not consist of the system's properties such as, masses, moment of inertia, etc... Thus, kinematic models are not sufficient to control the spherical rolling robots.

In practice, the system behavior is affected by unmodeled dynamics, parameter variations, uncertainties and disturbances. For instance, when the spherical rolling robot rolls on a surface, the static and the dynamic friction cannot be neglected. Since conventional controllers are time-invariant controllers, these terms cause discontinuities and nonlinearities which render conventional control invalid. Therefore, advanced intelligent control techniques are needed to overcome the stated shortcomings.  The control scheme used in this paper is called feedback-error-learning and was firstly proposed in \cite{Gomi} for the control of robot manipulators. It is based on the parallel work of a fuzzy neural network (FNN)-based controller and a conventional feedback controller.

All systems are nonlinear and continuous in nature but, the data flow between receiver and transmitter occurs in discrete-time domain in practice resulting in some modeling errors. In \cite{Munoz}, the modeling errors are estimated by using an artificial neural network (ANN) estimator, and tracking errors are reduced in discrete nonlinear systems. Even if the dynamic equations for the system at hand would be available, uncertainties can exist because of the noise on the sensors, environmental changes and/or nonlinear characteristics of the actuators. Radial basis function neural networks with an online learning algorithm have been proposed to handle the uncertainty problem \cite{Huang}. The dynamics of mobile robots with uncertainties and external disturbances have been studied in \cite{Das}. Self-recurrent wavelet neural networks are also proposed to estimate model uncertainties and external disturbances in the dynamics of nonholonomic wheeled mobile robots \cite{Park}. Dual neuro-adaptive control for the discrete-time dynamic control of non-holonomic mobile robots is introduced in \cite{Bugeja} which results in a major improvement in tracking performance, despite the plant uncertainty and unmodeled dynamics.

As a solution to the modeling errors and uncertainties problems mentioned above, an adaptive FNN-based controller with sliding mode control (SMC) theory-based learning algorithm is proposed for the control of spherical rolling robots. FNNs are a fusion of the capability of fuzzy reasoning to handle uncertain information and the capability of ANNs to learn from input-output data sets. Thus, FNNs are preferable approaches in engineering fields \cite{Topalov2,Kayacan2011,Khanesar}. Various learning algorithms have been proposed for FNNs. The gradient-based algorithm works well when the system at hand has very slow variations in its dynamics. However, since the gradient-based algorithms (e.g. dynamic back propagation) include partial derivatives, the convergence speed may be slow. Especially when the search space is complex. Moreover, the tuning process can easily be trapped into a local minimum \cite{Venelinov_1}. To alleviate the problems mentioned, the use of evolutionary approaches has been suggested \cite{Zhou}. However, the stability of such approaches is questionable and the optimal values for the stochastic operators are difficult to derive. Furthermore, the computational burden is very high. To overcome these issues, a novel SMC theory-based algorithm is proposed in this study. The system can provide adaptation to parameter variations, uncertainties and disturbances through the SMC theory-based online learning algorithm and also shows robust behaviour through the nature of SMC theory. SMC theory-based learning algorithms cannot only make the overall system more robust, but also ensure faster convergence than the traditional learning techniques in online tuning of ANNs and FNNs \cite{Efe2000}. The main contributions of this study beyond the state of the art are a novel SMC theory-based online learning algorithm for FNNs and the use of the proposed control algorithm in the velocity control of spherical rolling robots. To the best knowledge of the authors, this is the first time such an approach is ever used for the velocity control of a spherical rolling robot.

This paper is organized as follows: The mathematical model of the spherical rolling robot is presented in Section II. In Section III, the adaptive neuro-fuzzy control approach is used to design an intelligent controller, and the basics of the SMC theory-based online learning algorithm are given. Simulation results and comparisons are presented in Section IV. Finally, a brief conclusion of the study is given in Section V.

\section{The Mathematical Model of a Spherical Rolling Robot}\label{Modeling of Spherical Rolling Motion}

\begin{figure}[hbt]
\begin{center}
  \includegraphics[width=3.4 in]{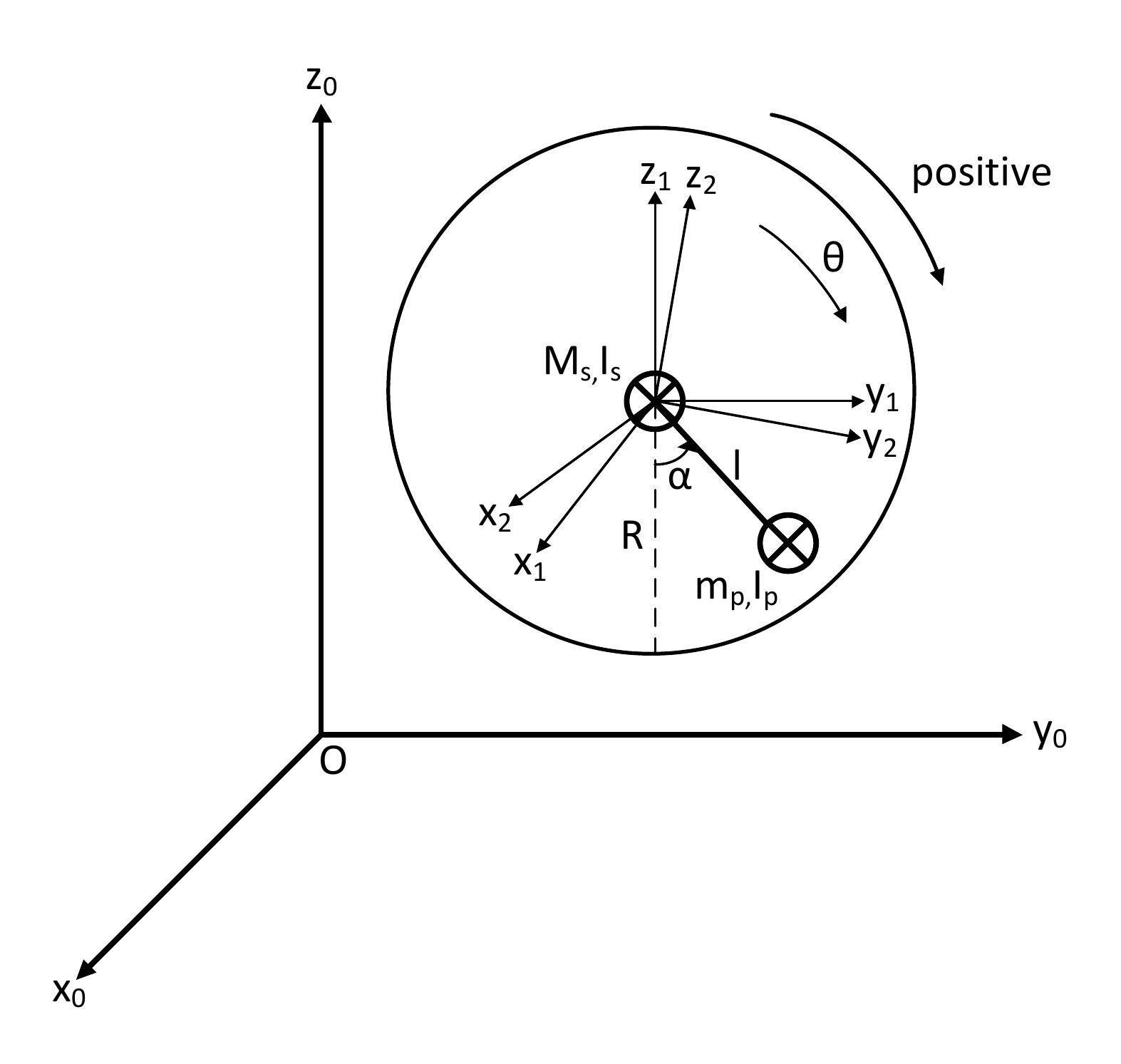}
  \caption{Modeling of rolling motion about transversal axis for overall translation along $O-y$}\label{sphere}
\end{center}
\end{figure}

The mathematical model of the spherical rolling robot is similar to the one presented by Kayacan et. al. \cite{Erkan}. The main difference is that viscous friction has been added into the equations of motion. A schematic illustration of the spherical rolling robot is given in Fig. \ref{sphere}.

\subsection{Kinematic Model}

Formulation of the kinematic and dynamic equations governing the motion of the spherical rolling mechanism with a pendulum is based on the following assumptions:

\begin{enumerate}
  \item The sphere rolls over a perfectly horizontal surface without slipping.
  \item The center of mass of the entire system is also the geometric center of the spherical shell.
  \item The pendulum is in vertical downward position when the sphere is in static equilibrium.
\end{enumerate}

A schematic illustration of the spherical rolling robot is given in Fig. \ref{sphere}. $R_{f_{0}}$, $R_{f_{1}}$ and $R_{f_{2}}$ denote respectively, the reference frames $O_{0}-X_{0}Y_{0}Z_{0}$, $O_{1}-X_{1}Y_{1}Z_{1}$ and $O_{2}-X_{2}Y_{2}Z_{2}$. $R_{f_{0}}$ represents the inertial reference frame fixed to the motion ground. $R_{f_{1}}$ is a moving frame attached to the center of the sphere and allowed to translate only with respect to $R_{f_{0}}$. $R_{f_{2}}$ is another moving frame attached also to the center of the sphere but allowed to rotate only with respect to $R_{f_{1}}$ . Relative angular positions between these frames can be described by several methods such as Euler angles, Tait-Bryan angles, Roll-Pitch-Yaw angles, etc.

The variables of the spherical rolling robot are represented in Table \ref{kinematic}.

\begin{table}[h!]
\caption{NOMENCLATURE}
\centering
\begin{tabular}{c c}
\hline
$\theta$ & Rolling angle of the sphere around the x axis\\
$\alpha$ & Degree of freedom of pendulum \\
R & Radius of the sphere \\
l & Distance between the center of the sphere and \\
 & the center of the pendulum\\
g & Gravitational acceleration \\
$\zeta$ & Damping coefficient \\[1ex]
\hline
\end{tabular}
\label{kinematic}
\end{table}

The angular velocity vector $\mathbf{\omega_{s}}$ and linear velocity vector $\mathbf{v_{s}}$ of the center of the sphere, the position vector $\mathbf{r_{p_{0}}}$, the angular velocity vector $\mathbf{\omega_{p}}$ and the linear velocity vector $\mathbf{v}_{p}$ of the mass center of the pendulum are given by:
\begin{eqnarray}
\mathbf{\omega_{s}} & = & - \dot{\theta} \mathbf{i} \\
\mathbf{v_{s}} & = & -R \dot{\theta} \mathbf{j}\\
\mathbf{r_{p}} & = & l \sin{(\alpha-\theta)} \mathbf{j} -l \cos{(\alpha-\theta)} \mathbf{k} \\
\mathbf{\omega_{p}} & = & (\dot{\alpha}-\dot{\theta})\mathbf{i} \\
\mathbf{v}_{p} & = & \big(-R \dot{\theta} + ( \dot{\alpha}-\dot{\theta}) l \cos{(\alpha-\theta)} \big) \mathbf{j} \nonumber \\
               & & +  \big( ( \dot{\alpha}-\dot{\theta}) l \sin{(\alpha-\theta)} \big) \mathbf{k}
\end{eqnarray}
where $\theta$, $\alpha$, $R$ and $l$  represent the rolling angle of the sphere around the x axis, the rotation of the pendulum around the x axis, the radius of the sphere and the distance between the center of the sphere and the center of the pendulum, respectively. Similarly, $\mathbf{i}$, $\mathbf{j}$ and $\mathbf{k}$ represent the unit vector on the x, y and z axes, respectively. It is to be noted that x,y and z axes denote the inertial reference frame fixed to the motion ground.

\subsection{Dynamic Model}

Let $E_{k}$ and $E_{p}$ denote respectively the total kinetic and potential energy of the system. With $M_{s}$ and $m_{p}$ representing the masses of the sphere and pendulum, $I_{s}$ and $I_{p}$ representing the mass moment of inertia, and $v_{s}$, $\omega_{s}$, $v_{p}$ and $\omega_{p}$ representing the linear and angular velocities of the sphere and pendulum, $r_{p-z}$ representing the vertical position of the mass center of the pendulum, the Lagrangian function $L$ including only the terms due to rotations around the transversal axis is then written as follows:

\begin{eqnarray}
L & =  & E_{k}-E_{p} \nonumber \\
 & =  & \frac{1}{2}M_s\|\mathbf{v_s}\|^2+\frac{1}{2}I_{s}\|\mathbf{\omega_s}\|^2+\frac{1}{2}m_p\|\mathbf{v_p}\|^2 +\frac{1}{2}I_{p}\|\mathbf{\omega_p}\|^2 \nonumber \\
&& - m_{p}gr_{p-z} \nonumber \\
 & =  & \frac{1}{2}M_s(-R \dot{\theta})^2 + \frac{1}{2}I_{s}(-\dot{\theta})^2 + \frac{1}{2}I_{p}(\dot{\alpha}-\dot{\theta})^2 \nonumber \\
&& + \frac{1}{2}m_p \Big( \big(-R \dot{\theta} + (\dot{\alpha}-\dot{\theta}) l \cos{(\alpha-\theta)}\big)^2 \nonumber \\
&&+ \big((\dot{\alpha}-\dot{\theta}) l \sin{(\alpha-\theta)}\big)^2\Big) - m_{p}g l \cos{(\alpha-\theta)}
\end{eqnarray}

It is assumed that the viscous friction operates between the sphere and the surface. The loss due to the viscous friction is written in an energy dissipation function that depends on the velocities of the system and the damping constant:

\begin{equation}\label{sfunction}
    S = \frac{1}{2}\zeta\dot{q}_i^2 =  \frac{1}{2}\zeta(\dot{\theta}^2+\dot{\alpha}^2)
\end{equation}

For translation along $O-y$, the Euler-Lagrange equations of the system are written as follows:

\begin{equation}\label{lagrange}
\frac{d}{dt}(\frac{\partial L}{\partial \dot{q}_i})-\frac{\partial L}{\partial q_i}+\frac{\partial S}{\partial \dot{q}_i}=Q_i
\end{equation}
where $q_{1}= \theta $ and $q_{2}= \alpha $ are the generalized coordinates. In fact, when the pendulum is rotated through an input torque, a reaction torque about the shaft occurs in the opposite direction \cite{Ming}. $Q_1=Q_2=\tau$ represents the input torque to rotate the pendulum.

\begin{eqnarray}
  \begin{array}{c}
   Q_\theta=\tau \\
   Q_\alpha=\tau \\
  \end{array}
\end{eqnarray}

As the equations of motion of a mechanical system can be written as follows:
\begin{equation}
M\big(q(t)\big)\ddot{q(t)}+C\big(q(t),\dot{q(t)}\big)+G\big(q(t)\big)=u(t)
\end{equation}

The equations of motion can be finally written in the following matrix form:
\begin{eqnarray}\label{sondenklem}
\left[
  \begin{array}{cc}
   M_{11} & M_{12}\\
   M_{21} & M_{22} \\
  \end{array}
  \right]
\left[
  \begin{array}{c}
   \ddot {\theta}\\
   \ddot {\alpha} \\
  \end{array}
  \right]
+
\left[
  \begin{array}{c}
   C_{11} \\
   C_{21} \\
  \end{array}
  \right]
+
  \left[
  \begin{array}{c}
   G_{11} \\
   G_{21} \\
  \end{array}
   \right]
  =
\left[
  \begin{array}{c}
   \tau \\
   \tau \\
  \end{array}
  \right]
\end{eqnarray}
where
\begin{eqnarray}
M_{11} & = & M_s R^2 + m_p R^2 + m_p l^2 + I_s + I_p  \nonumber\\
       &   & + 2 m_p R l \cos{(\alpha-\theta)} \nonumber\\
M_{12} & = & M_{21} = - m_p l^2 - I_p - m_p R l \cos{(\alpha-\theta)}\nonumber \\
M_{22} & = & m_p l^2 + I_p \nonumber \\
C_{11} & = & m_p R l \sin{(\alpha-\theta)} (\dot {\alpha}-\dot {\theta})^2 + \zeta \dot {\theta} \nonumber \\
C_{21} & = & \zeta \dot {\alpha} \nonumber \\
G_{11} & = & - G_{21}  = - m_p g l \sin{(\alpha-\theta)}  \nonumber
\end{eqnarray}

In this study, velocity control of the spherical rolling robot is studied by using the above equations of motion. If rolling motion over curvilinear trajectories is desired, the equations of motion presented in \cite{Erkan} can be used. Different from the approach in \cite{Erkan}, a viscous friction term is added in this study.

\section{The Adaptive Neuro-Fuzzy Control Approach}\label{The Adaptive Neuro-fuzzy Control Approach}

\subsection{The Control Scheme and the Adaptive Neuro-Fuzzy Inference System}

In real-time applications, structural and parametric uncertainties such as unmodeled dynamics and physical disturbances cause unwanted effects on the system behaviour. Like tire dynamics, the spherical rolling robot dynamics are affected by the slip ratio and variations of the surface conditions. Thus, spherical rolling robots should have an adaptive intelligent controller with respect to the high uncertainty and the variation in the surface properties. While conventional controllers struggle with these problems, neuro-fuzzy control can overcome these limitations and provide higher robustness.

In the presented approach, a conventional PD or PID controller and a neuro-fuzzy controller are working in parallel. The conventional controller is used to guarantee the global asymptotic stability of the system in compact space. The PD control law is written as follows:
\begin{equation}  \label{PD_control_law}
\tau_{c}=k_{p}e+k_{d}\dot{e} \\
\end{equation}
where $e=\theta_d-\theta$ is the feedback error, $\theta_d$ is the desired position value, $k_p$ and $k_d$ are the controller gains.

\begin{figure}[hbt]
\begin{center}
\includegraphics[width=3.6 in]{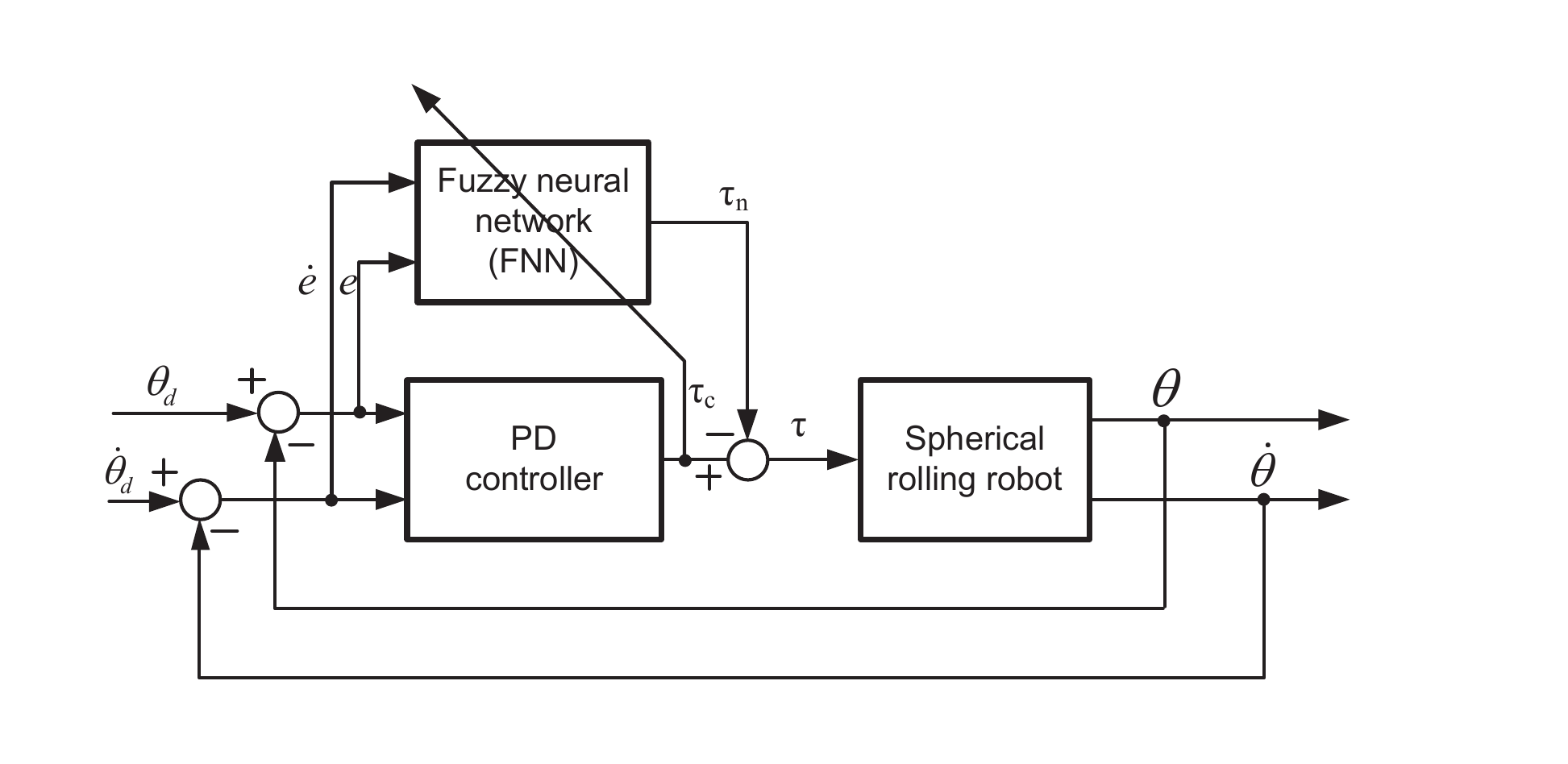}
\caption{Block diagram of the proposed control scheme for the PD controller case}\label{blockdiagram}
\end{center}
\end{figure}

\begin{figure}[hbt]
\begin{center}
\includegraphics[width=3.6 in]{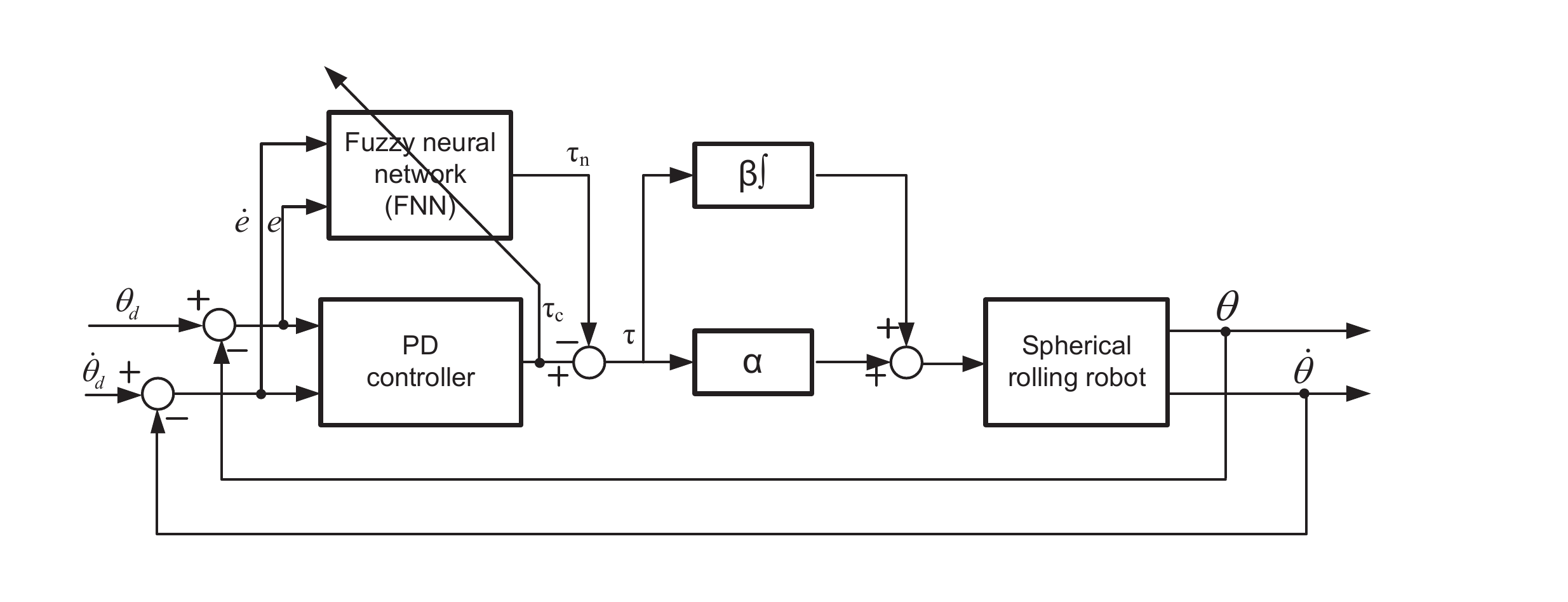}
\caption{Block diagram of the proposed control scheme for the PID controller case}\label{blockdiagramPID}
\end{center}
\end{figure}

\begin{figure}[!t]
\begin{center}
\includegraphics[width=3.6 in]{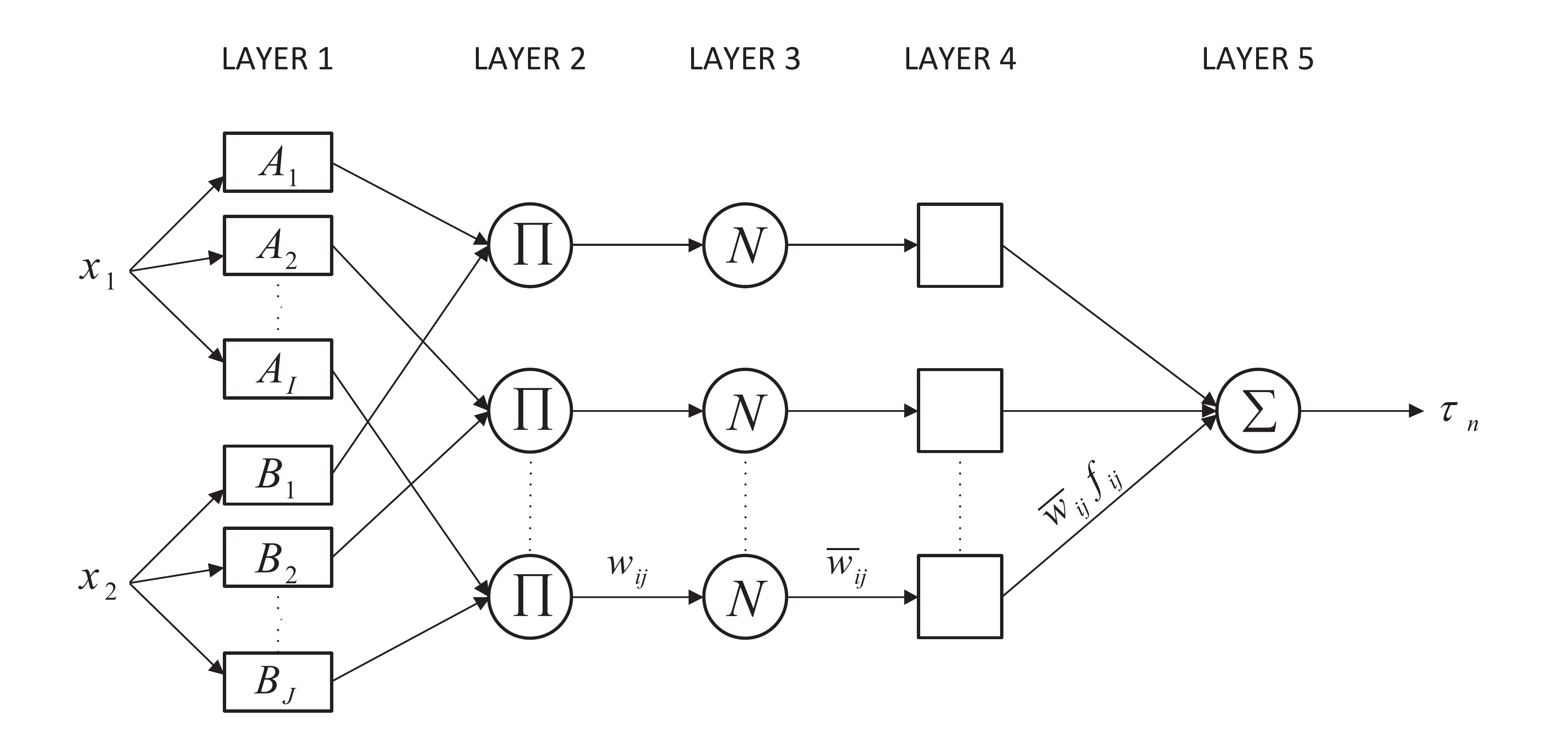}
\caption{The FNN}\label{neurofuzzy}
\end{center}
\end{figure}

The proposed control scheme for the PD controller case, the PID controller case and the structure of the proposed FNN are illustrated in Fig. \ref{blockdiagram}, Fig. \ref{blockdiagramPID} and Fig. \ref{neurofuzzy}, respectively. The terms of $\alpha$ and $\beta$ are the coefficients of the PI controller. As can be seen from the proposed control scheme, a FNN which has two inputs $x_1(t)=e(t)$, $x_2(t)=\dot{e}(t)$ and one output $f_{ij}$, is applied as a feedback controller.

The fuzzy \emph{If-Then} rule of a zeroth-order Takagi-Sugeno-Kang (TSK) model with two input variables where the consequent part is a linear function of the input variables can be defined as follows:

\begin{equation}
R_{ij}: \;\; \textrm{If} \; x_1 \; \textrm{is} \;\; A_i \;\; \textrm{and} \; x_2 \; \textrm{is} \;\; B_j, \;\; \textrm{Then} \; f_{ij}=d_{ij}
\end{equation}
where $x_1$ and $x_2$ are the inputs of the FNN, $A_i$ and $B_j$ are fuzzy sets corresponding to the input 1 and input 2, respectively. The zeroth-order function, $f_{ij}$  is the consequent part of the rules where $I$ $(i=1,\dots,I)$  and $J$ $(j=1,\dots,J)$ being the number of membership functions used for the input 1 and input 2, respectively.

The following are the basic explanations for Fig. \ref{neurofuzzy}:

  \textbullet Layer 1: The first layer is called input later. This layer maps the crisp inputs $x_1$ and $x_2$ into fuzzified values using membership functions. Every node has a node function described as follows:

\begin{eqnarray}
      O^{1} _{i} =\mu_{A_i}(x_1)  \nonumber \\
      O^{1} _{j} = \mu_{B_j}(x_2)
\end{eqnarray}
where $\mu_{A_i}(x_1)$ and $\mu_{B_j}(x_2)$ are the membership values for the inputs 1 and 2, respectively. In the case of Gaussian membership functions, $\mu_{A_i}(x_1)$ and $\mu_{B_j}(x_2)$ are written as follows:
\begin{eqnarray}\label{mu}
\mu_{A_i}(x_1) =\exp \Big(-(\frac{x_1 - c_{A_i}}{\sigma _{A_i}})^2 \Big)   \nonumber \\
\mu_{B_j}(x_2) =\exp \Big(-(\frac{x_2 - c_{B_j}}{\sigma _{B_j}})^2 \Big)
\end{eqnarray}
where $c$ and $\sigma$ are respectively the mean and the standard deviation of the membership functions. These parameters $c, \sigma >  0$ are the tunable parameters of the neuro-fuzzy structure.

  \textbullet Layer 2: The strength of the rule is obtained in this layer. The output of each node, which calculates the firing strength $w_{ij}$ of a rule, is multiplied with all incoming signals. The node function of this layer is written as follows:
\begin{equation}\label{layerO21}
      O^{2} _{ij} = w_{ij} = \mu_{A_i}(x_1) \mu_{B_j}(x_2)
\end{equation}

The previous equation (\ref{layerO21}) is rewritten considering (\ref{mu}) as follows:
\begin{equation}\label{layerO22}
      O^{2} _{ij} = w_{ij} = \exp \Big(-(\frac{x_1 - c_{A_i}}{\sigma _{A_i}})^2 -(\frac{x_2 - c_{B_j}}{\sigma _{B_j}})^2 \Big)
\end{equation}

  \textbullet Layer 3: The normalized firing strength of each node is written as follows:
\begin{equation}\label{layerO3}
      O^{3} _{ij} = \overline{w}_{ij} = \frac{w_{ij}}{\sum\limits_{i=1}^{I}\sum\limits_{j=1}^{J}{w}_{ij}}
\end{equation}

  \textbullet Layer 4: Each node in this layer is an adaptive node with node function written by the following equation:
\begin{equation}\label{layerO4}
      O^{4} _{ij} = \overline{w}_{ij}f_{ij}
\end{equation}

  \textbullet Layer 5: There is one node where all incoming signals are summed in this layer. The output signal of the neuro-fuzzy network $\tau_n$ is calculated by the function described as follows:
\begin{equation}\label{layerO5}
      O^{5} =  \tau_n(t) = \sum\limits_{i=1}^{I} \sum\limits_{j=1}^{J} \overline{w}_{ij}f_{ij} =  \sum\limits_{i=1}^{I} \sum\limits_{j=1}^{J} \overline{w}_{ij} d_{ij}
\end{equation}

The control input fed to the system is the overall torque  $\tau$  determined as follows:
\begin{equation} \label{GrindEQ__15_}
\tau =\tau_{c} -\tau_{n}
\end{equation}
where $\tau_{c}$ and $\tau_{n}$  are the input torques generated by the PD controller and the neuro-fuzzy feedback controller, respectively.

\subsection{The Sliding Mode Learning Algorithm}

The input signals and their time derivatives $x_1 (t), x_2 (t), \dot{x}_1 (t), \dot{x}_2 (t)$ cannot have infinite values. Thus, they are bounded as follows:
\begin{eqnarray}
\mid x_1 (t) \mid \le B_x, \; \; \mid x_2 (t) \mid \le B_x \quad \forall	t\nonumber \\
\mid \dot{x}_1 (t)  \mid \le B_{\dot{x}}, \; \; \mid \dot{x}_2 (t)  \mid \le B_{\dot{x}} \quad \forall	t \nonumber \\
\end{eqnarray}

Similarly, the input torque to the system and its time derivative are bounded:
\begin{equation}
\mid \tau(t) \mid \le B_{\tau}, \; \; \mid \dot{\tau} (t) \mid \le B_{\dot{\tau}}  \quad \forall	 t
\end{equation}

The output of the PD controller $\tau _{c} \left(t\right)$ is defined as a time-varying sliding surface by using the principles of SMC theory \cite{Utkin}:
\begin{equation}\label{Sc}
S_{c} \left(\tau _{n} ,\tau \right)=\tau _{c} \left(t\right)=\tau _{n} \left(t\right)+\tau \left(t\right)=0
\end{equation}

The sliding surface $S_{p} \left(e,\dot{e}\right)$ is defined as follows:

\begin{equation}
S_{p} \left(e,\dot{e}\right)=\dot{e} + \lambda e
\end{equation}
where $\lambda$ is a constant determining the slope of the sliding surface.

\textit{Definition:} A sliding motion exists on the sliding surface $S_{c}\left(\tau_{n},\tau\right)=\tau_{c}\left(t\right)=0$ after time $t_{h} $, if the condition $S_{c} (t)\dot{S}_{c} (t)=\tau _{c} \left(t\right)\dot{\tau }_{c} \left(t\right)<0$ is satisfied for all $t$ in some nontrivial semi-open subinterval of time of the form $\left[t,t_{h} \right)\subset \left(-\infty ,t_{h} \right)$.

Since it is desired to design a dynamical feedback adaptation mechanism, or online learning algorithm for the FNN parameters, the sliding mode condition of the above definition is applied.

\newtheorem{theorem}{Theorem}
\begin{theorem}

The adaptation laws for the parameters of the proposed FNN are given by the following equations:

\begin{equation} \label{eq1}
\dot{c}_{A_{i} } = \dot{x_{1}} + s_{A_i} \alpha \textrm{sgn}\left(\tau _{c} \right)
\end{equation}

\begin{equation}\label{eq2}
\dot{c}_{B_{j} } =\dot{x_{2}} + s_{B_j} \alpha \textrm{sgn}\left(\tau _{c} \right)
\end{equation}

\begin{equation}\label{eq3}
\dot{\sigma }_{A_{i} } = - \bigg( \sigma _{A_{i}} + \frac{ (\sigma_{A_{i}} )^3}{s_{A}^{T} s_{A} } \bigg) \alpha  \textrm{sgn}\left(\tau _{c} \right)
\end{equation}

\begin{equation}\label{eq4}
\dot{\sigma }_{B_{j} } = - \bigg( \sigma _{B_{j}} + \frac{ (\sigma_{B_{j}} )^3 }{s_{B}^{T} s_{B} } \bigg) \alpha  \textrm{sgn}\left(\tau _{c} \right)
\end{equation}

\begin{equation}\label{eq5}
\dot{f}_{ij} =-\frac{\overline{W}_{ij} }{\overline{W}^{T} \overline{W}} \alpha \textrm{sgn}\left(\tau _{c} \right)
\end{equation}

\end{theorem}

The vectors in (\ref{eq1})-(\ref{eq4}) are defined as follows:
\begin{eqnarray}
s_{A_i}=x_1-c_{A_i} \; \textrm{and} \; s_{A} =\left[s_{A_{1} } \, s_{A_{2} } ...\, s_{A_{I} } \right]^{T} \\
s_{B_j}=x_2-c_{B_j} \; \textrm{and} \; s_{B} =\left[s_{B_{1} } \, s_{B_{2} } ...\, s_{B_{J} } \right]^{T}
\end{eqnarray}
where $\alpha$ is the learning rate which is a sufficiently large positive design constant which satisfies the following inequality:

\begin{equation}\label{alpha}
B_{\dot{\tau}} < \alpha
\end{equation}

This ensures that for a given arbitrary initial condition $\tau _{c}(0)$, the learning error $\tau _{c}(t)$ will converge to zero within finite time $t_{h}$.

\emph{Proof:} The reader is referred to Appendix A.

If $\lambda$ is considered $\lambda=\frac{k_{p}}{k_{d}}$, the relation between the sliding line $S_{p}$ and the zero adaptive learning error level $S_{c}$ can be written as follows:

\begin{equation}
S_{c} =\tau _{c} =k_{d} \dot{e}+k_{p} e=k_{d} \left(\dot{e}+\frac{k_{p} }{k_{d} } e\right)=k_{d} S_{p}
\end{equation}

The tracking performance of the velocity control system in the spherical rolling robot is analyzed by the following Lyapunov function candidate:

\begin{equation}\label{lyapunov}
V_{p} =\frac{1}{2} S_{p}^{2}
\end{equation}

\newtheorem{theorem2}{Theorem2}
\begin{theorem}
If the adaptation strategy for the adjustable parameters of the FNN is chosen as in (\ref{eq1})-(\ref{eq5}), then the negative definiteness of the time derivative of the Lyapunov function in (\ref{lyapunov}) is ensured.
\end{theorem}

\emph{Proof:} The reader is referred to Appendix B.

\section{Simulation Results and Discussion}

The numerical values used in \cite{Javadiasme,Joshi,Ming} are considered for the numerical values in this study which are $M_s=3$ $kg$, $m_p=2$ $kg$, $R=0.2$ $m$, $l=0.075$ $m$ and $g=9.81$ $m/s^2$. The damping coefficient, $\zeta$, in the equations of motion is set to $0.2$, and the sampling period of the simulations is set to $0.001$ $s$. The number of membership functions for the input $1$ and input $2$ is chosen as $I = J = 3$ for all the simulations.

SMC theory suffers from high-frequency oscillations called \emph{chattering} because of the input control law. Several approaches have been suggested to get rid of this problem. In this paper, the $sgn$ function in (\ref{eq1})-(\ref{eq5}) is replaced by the following equation to decrease the chattering effect:
\begin{equation}\label{sgn}
\textrm{sgn}\left(\tau _{c} \right) := \frac{\tau_c}{\mid \tau_c \mid +\delta}
\end{equation}
where $\delta=0.05$.

\subsection{Case 1: PD controller}

The coefficients of the PD controller are set to $k_p=1$ and $k_d=0.05$ by trial-and-error method. The following reference signal has been applied to the system (\ref{reference}):
\begin{equation}
\textrm{Reference}(t) = \left\{
\begin{array}{l l}
  1 \; \textrm{rad/s} & \quad \textrm{if $0<t\le5$ }\\
  2 \; \textrm{rad/s} & \quad \textrm{if $5<t\le10$ }\\
  1.5 \; \textrm{rad/s} & \quad \textrm{if $10<t\le15$ }\\
\end{array} \right.
\label{reference}
\end{equation}

\begin{figure}[b!]
\begin{center}
\includegraphics[width=3.4 in]{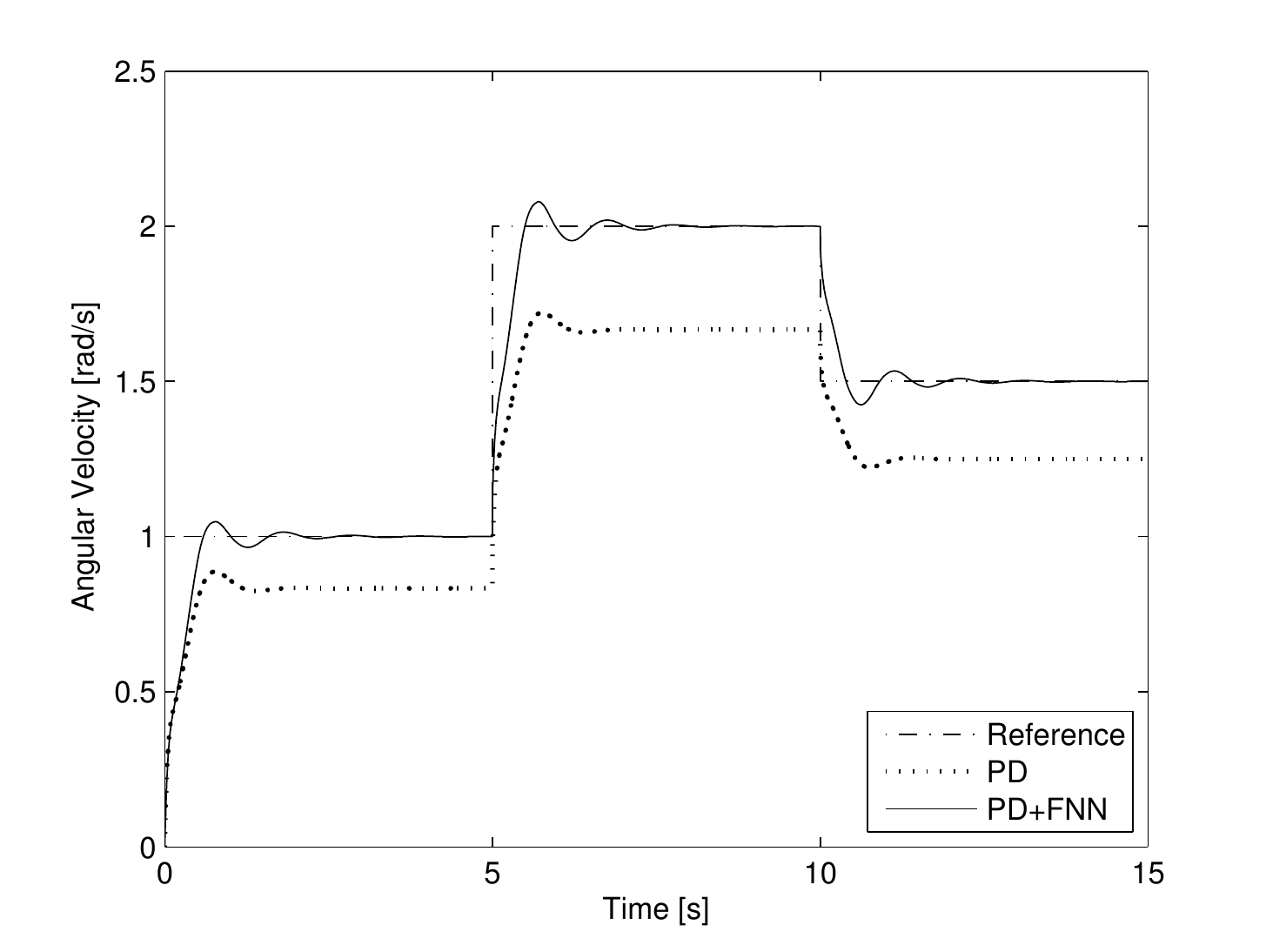}
\caption{The velocity response of the system for PD and PD+FNN controllers}\label{PDFNNsabitbhiz}
\end{center}
\end{figure}

\begin{figure}[h!]
\begin{center}
\includegraphics[width=3.4 in]{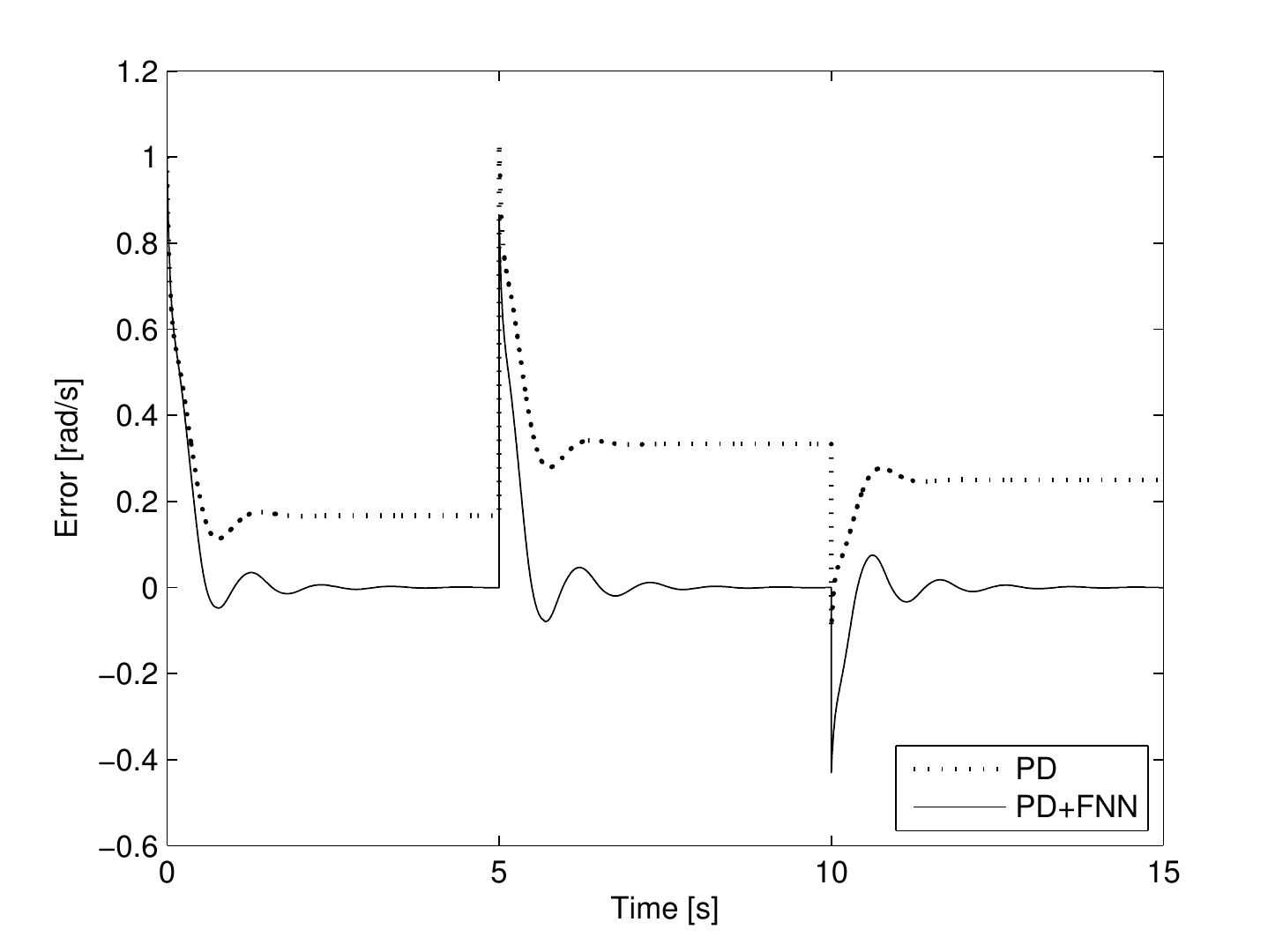}
\caption{The error of the system for PD and PD+FNN controllers}\label{errorhiz}
\end{center}
\end{figure}

Figures \ref{PDFNNsabitbhiz}-\ref{errorhiz} show the velocity response and the error of the system for the PD controller working alone and the PD controller working in parallel with the FNN, respectively. As can be seen from Fig. \ref{PDFNNsabitbhiz}, the PD controller cannot eliminate the steady state error for a time-varying step input. The system does not have any steady state error for the case of the FNN working in parallel with a PD controller.


\subsection{Case 2: PID controller}

The coefficients of the PID controller are set to $k_p=1, k_d=0.05, \alpha=1$ and $\beta=2$ by trial-and-error method. The reference signal for this simulation is the same the one (See \ref{reference}) in Case 1.

\begin{figure}[b!]
\begin{center}
\includegraphics[width=3.4 in]{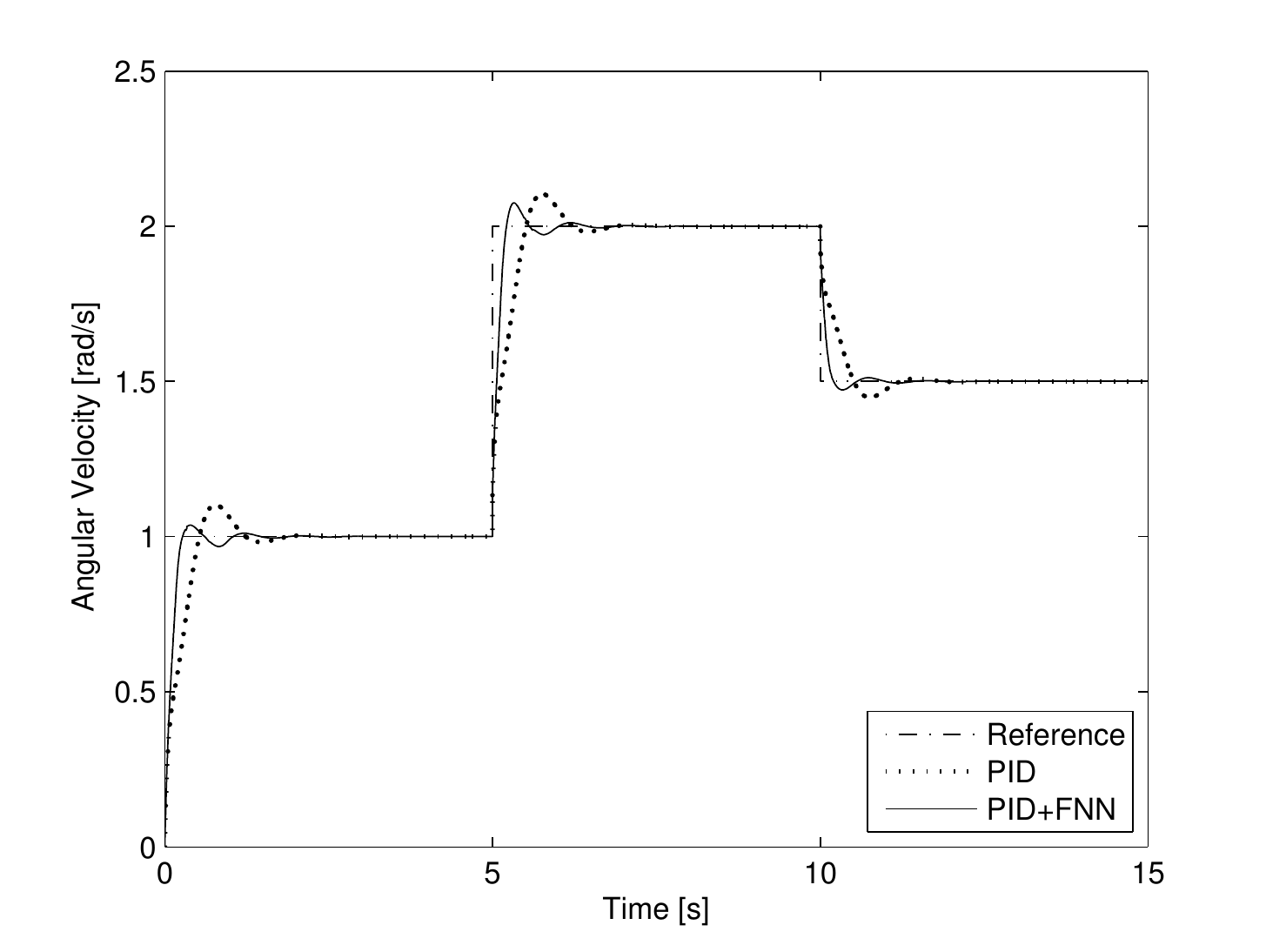}
\caption{The velocity response of the system for PID and PID+FNN controllers}\label{PIFNNsabitbhiz}
\end{center}
\end{figure}

\begin{figure}[h!]
\begin{center}
\includegraphics[width=3.4 in]{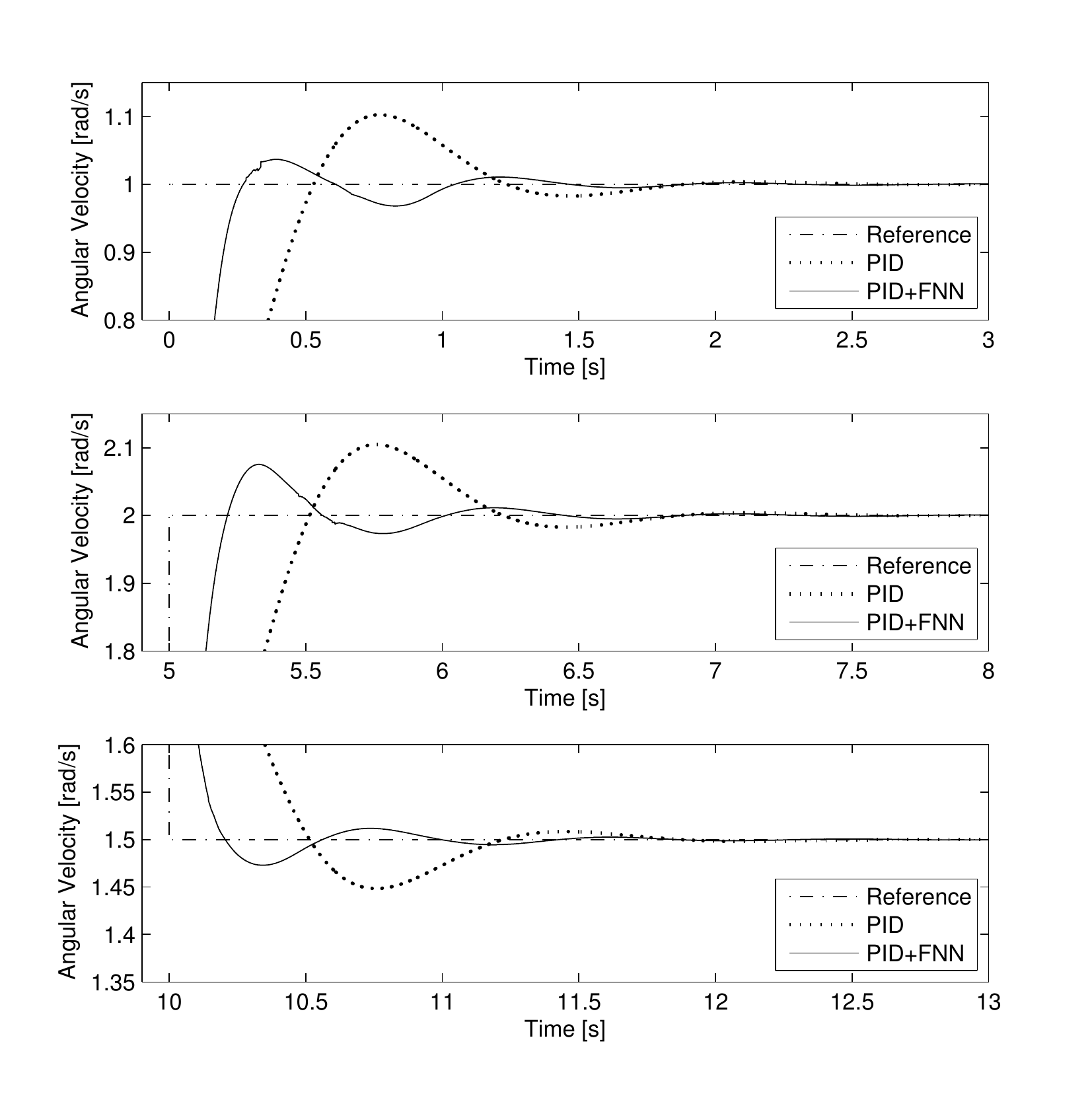}
\caption{Zoomed view of Figure \ref{PIFNNsabitbhiz}}\label{PIFNNsabitbhizzoom}
\end{center}
\end{figure}

In Figure \ref{PIFNNsabitbhiz} the responses of the PID controller and the combination of the FNN in parallel with this PID controller are illustrated. The FNN learns the system dynamics after a finite time duration, and results in a smaller rise time, overshoot and settling time. From Fig. \ref{PIFNNsabitbhiz} and \ref{PIFNNsabitbhizzoom} it can be concluded that the control scheme with the proposed SMC theory-based learning algorithm is able to not only eliminate the steady state error (shown in Case 1) but also to improve the transient response performance of the system (shown in Case 2). Figure \ref{PIFNNsabitbhizzoom} shows the zoomed view of Fig. \ref{PIFNNsabitbhiz} between $0^{th}-3^{rd}$, $5^{th}-8^{th}$ and $10^{th}-13^{th}$.


\begin{figure}[b!]
\begin{center}
\includegraphics[width=3.4 in]{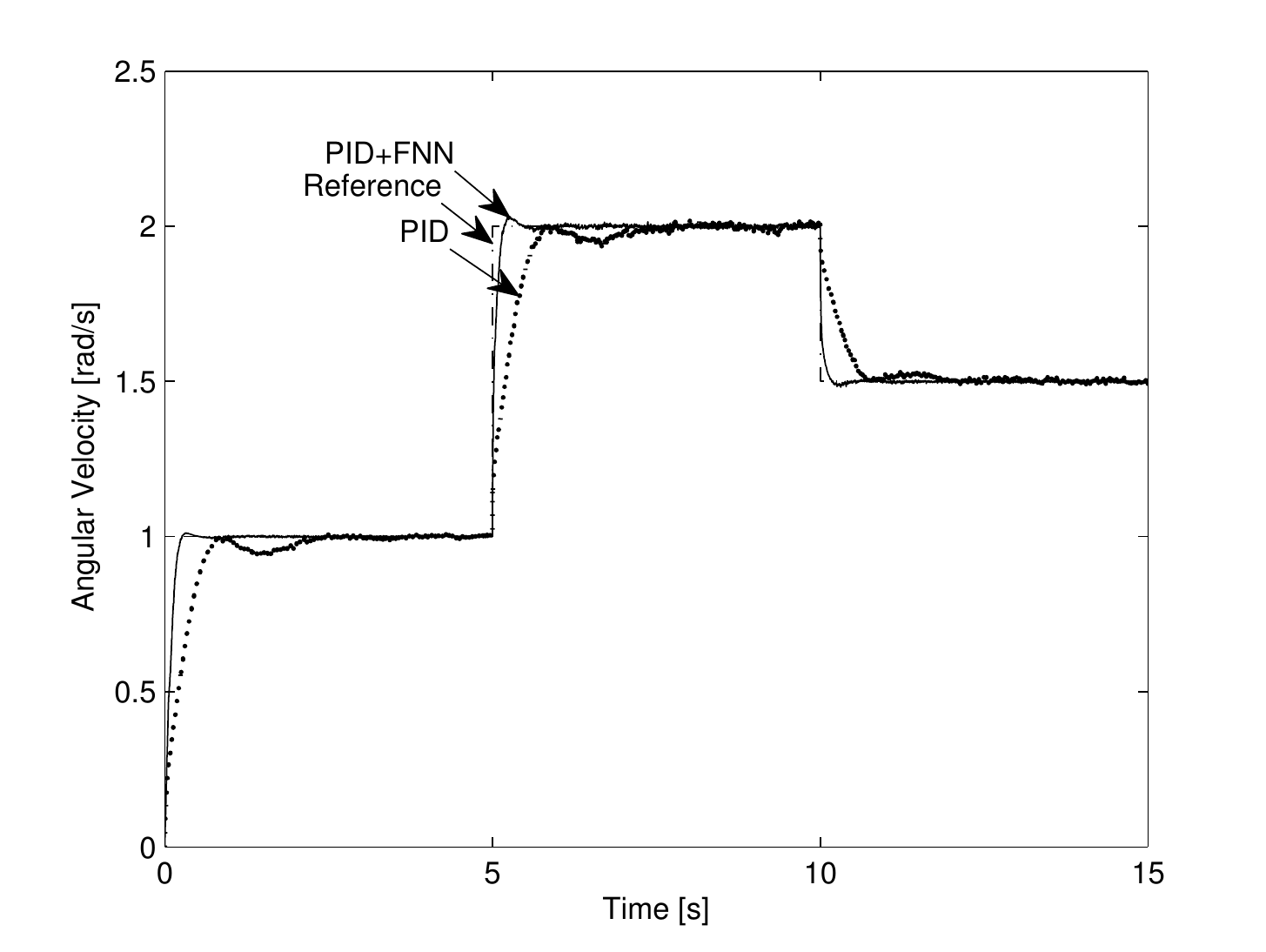}
\caption{The velocity response of the system for PID and PID+FNN controllers when the damping coefficient is set to $0.5$ with a noise level $SNR=20dB$}\label{PIFNNdegiskenbhiz}
\end{center}
\end{figure}

\begin{figure}[h!]
\begin{center}
\includegraphics[width=3.4 in]{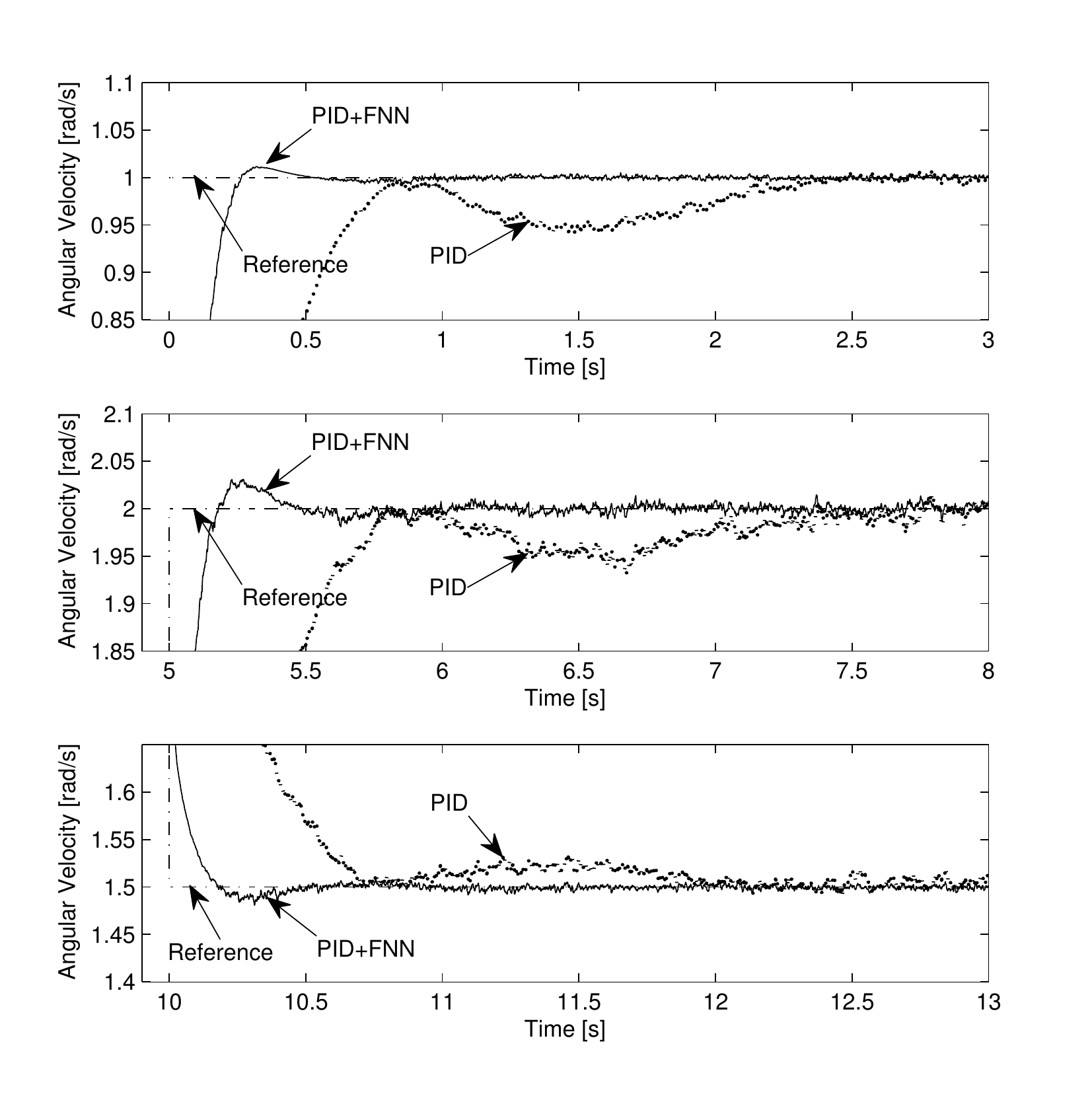}
\caption{Zoomed view of Figure \ref{PIFNNdegiskenbhiz}}\label{PDANFIShizbzoom}
\end{center}
\end{figure}

To test the robustness of this approach, the damping coefficient is set to $0.5$ with a noise level $SNR=20dB$. In Figure \ref{PIFNNdegiskenbhiz} the responses of the PID controller and the combination of the FNN working in parallel with the PID controller are illustrated. In Fig. \ref{PDANFIShizbzoom} close ups of Fig. \ref{PIFNNdegiskenbhiz} are shown. It can be observed that the adaptive neuro-fuzzy control approach is much more robust to the uncertainties and gives more satisfactory results regarding smaller rise time and settling time than the PID controller working alone.


In a next step, the following variation of the damping coefficient over time was applied to the system to test the robustness of the controllers:
\begin{equation}
\zeta (t) = \left\{
\begin{array}{l l}
  0.2 \; \textrm{rad/s} & \quad \textrm{if $0<t\le5$ }\\
  0.5 \; \textrm{rad/s} & \quad \textrm{if $5<t\le10$ }\\
  0.8 \; \textrm{rad/s} & \quad \textrm{if $10<t\le15$ }\\
\end{array} \right.
\label{zeta}
\end{equation}

\begin{figure}[h!]
\begin{center}
\includegraphics[width=3.4 in]{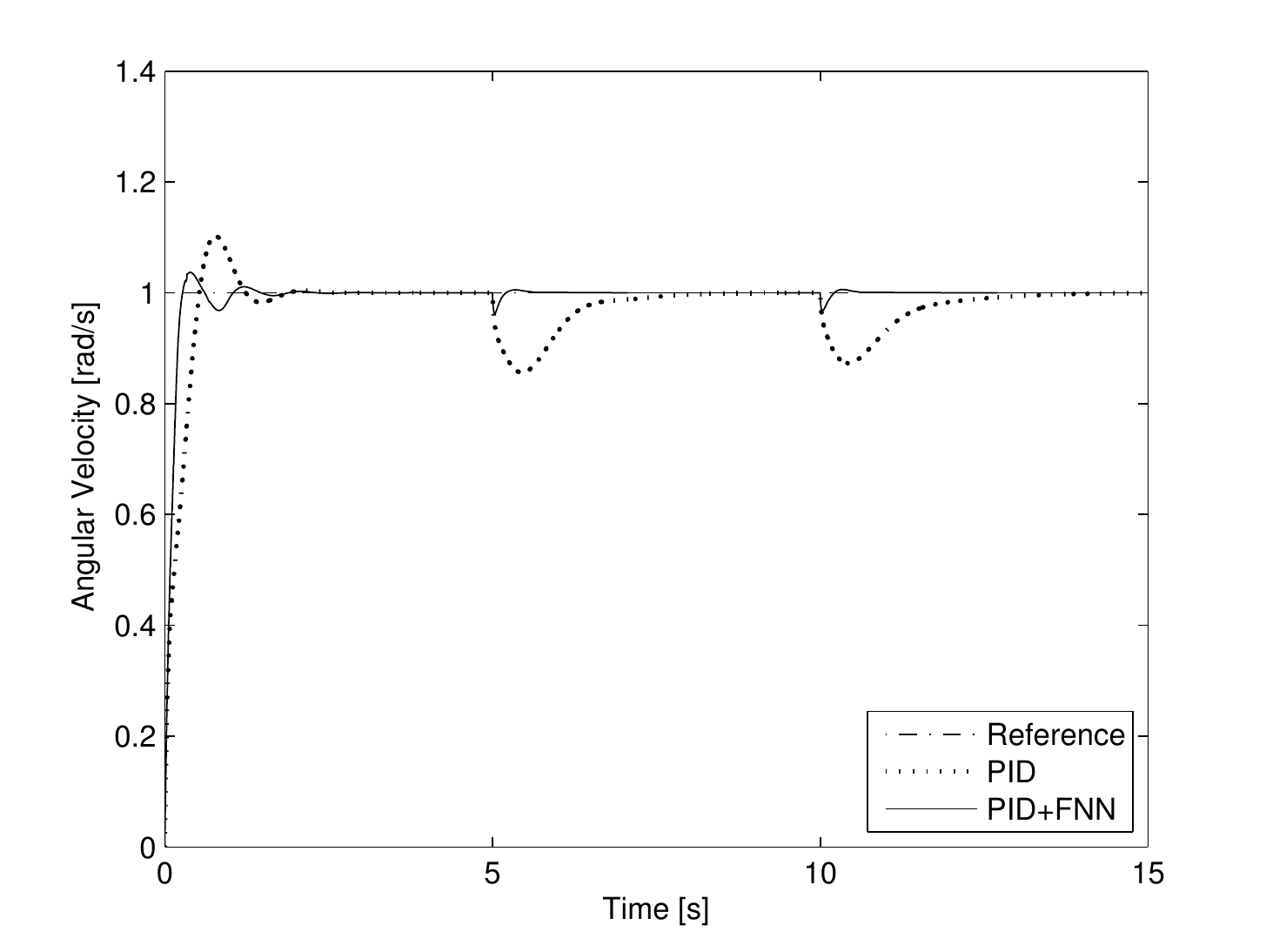}
\caption{The velocity response of the system for PID and PID+FNN controllers when the damping coefficient is set to $0.2$, $0.5$ and $0.8$ at $0^{th}$ second, $5^{th}$ second and $10^{th}$ second, respectively}\label{PIFNN3bhiz}
\end{center}
\end{figure}

\begin{figure}[h!]
\begin{center}
\includegraphics[width=3.4 in]{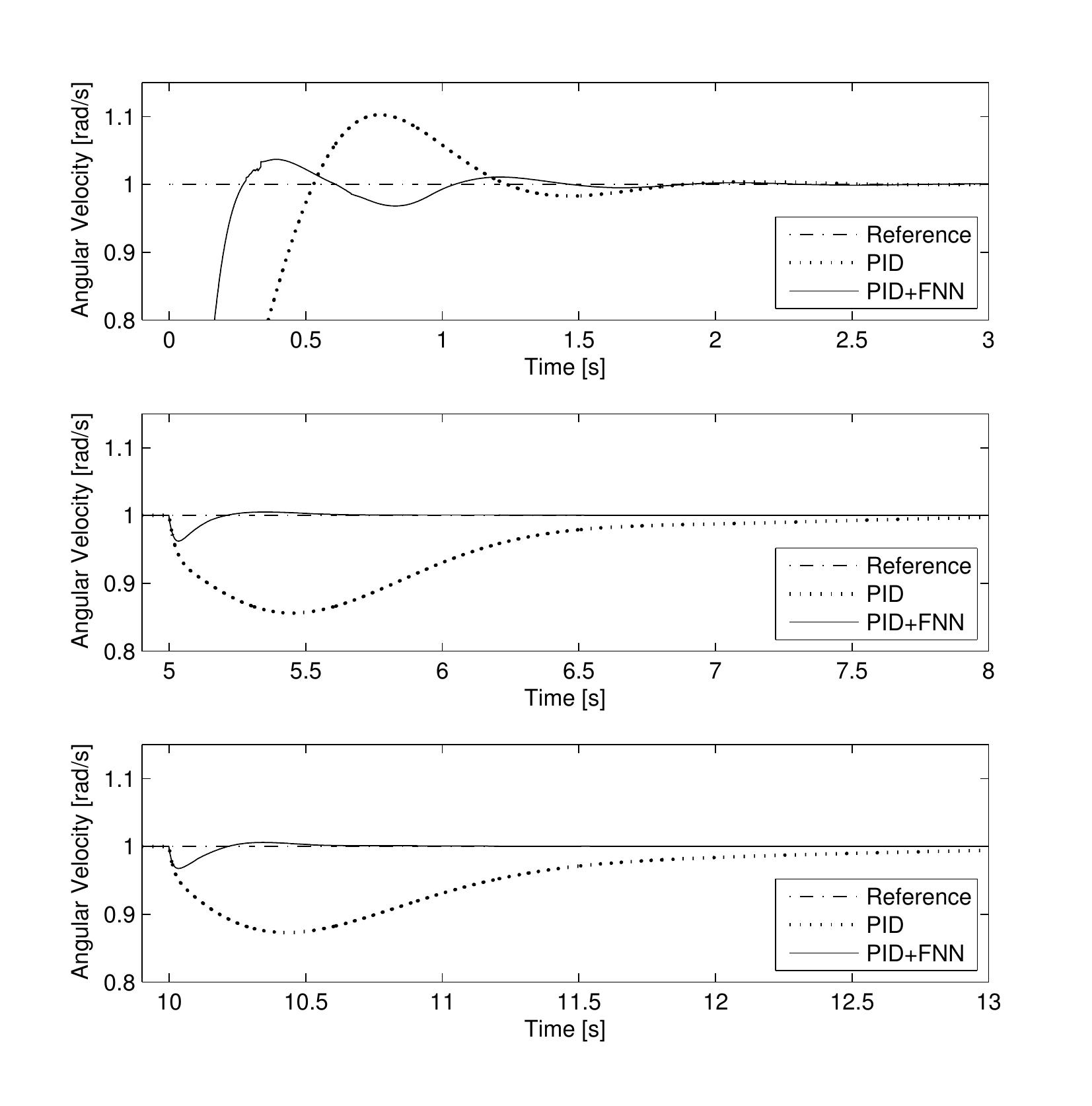}
\caption{Zoomed view of \ref{PIFNN3bhiz}}\label{PIFNN3bhizzoom}
\end{center}
\end{figure}

In Figure \ref{PIFNN3bhiz} the velocity responses of a conventional PID controller and the combination of the FNN working in parallel with this PID controller are shown, while close ups are shown in Fig. \ref{PIFNN3bhizzoom}. As can be seen from Fig. \ref{PIFNN3bhiz} - \ref{PIFNN3bhizzoom}, the FNN can adapt its parameters when the coefficient of viscous friction changes suddenly. For the case when the coefficient of the viscous friction changes, while the settling time for the PID controller working alone is approximately $2.5$ seconds, it is less than $1$ second for the FNN working in parallel with a PID controller. From these simulations it can be concluded that the adaptive neuro-fuzzy control approach is much more robust to the parameter variations compared to the case of the PID controller working alone.


\begin{figure}[h!]
\begin{center}
\includegraphics[width=3.4 in]{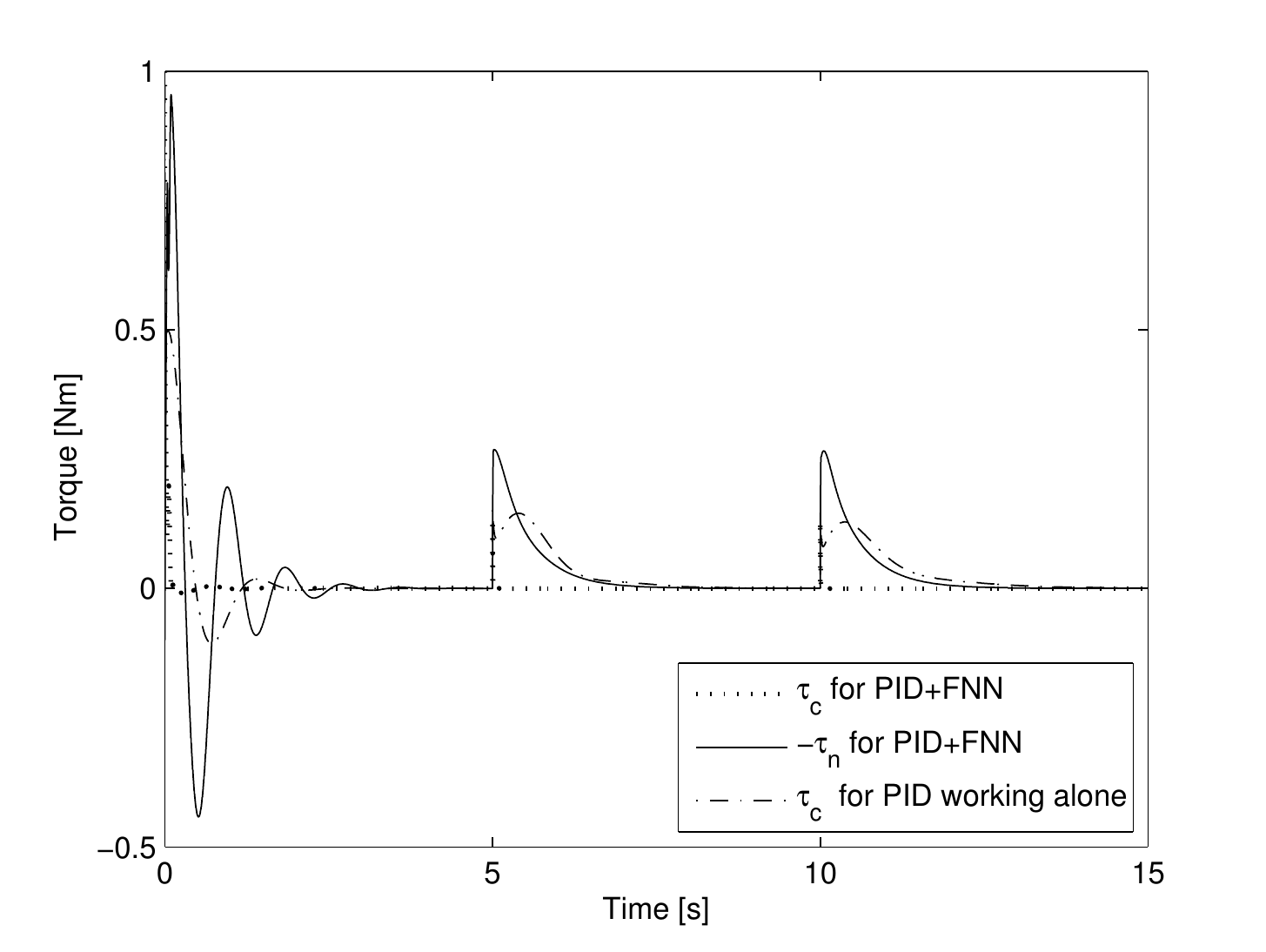}
\caption{The control signals coming from the PID controller and the FNN when the damping coefficient is set to $0.2$, $0.5$ and $0.8$ at $0^{th}$ second, $5^{th}$ second and $10^{th}$ second, respectively}\label{PIFNN3btorque}
\end{center}
\end{figure}

In Figure \ref{PIFNN3btorque} the control signals coming from the PID controller alone and the FNN working in parallel with the PID controller are illustrated for the case when the damping coefficient is set to $0.2$, $0.5$ and $0.8$ at $0^{th}$ second, $5^{th}$ second and $10^{th}$ second, respectively. In order not to confuse the reader, since the control input to the system is equal to $\tau_c - \tau_n$ when the PID controller works in parallel with the FNN, $\tau_n$ is multiplied by $-1$ in this figure. As can be seen from Fig. \ref{PIFNN3btorque}, the output of the PID controller is approximately zero while the output of the FNN changes as the damping coefficient changes as well. In other words, the FNN  takes the responsibility for controlling the system after a finite time duration.

\section{Conclusion}

In this study, an adaptive neuro-fuzzy controller with SMC theory-based online learning has been elaborated for velocity control of a spherical rolling robot subject to parameter variations and uncertainties in its dynamics. The simulation studies show that the proposed adaptive neuro-fuzzy control scheme (a conventional controller working in parallel with the FNN) results in a better performance and higher robustness than when compared to the conventional stand-alone controller. The proposed control algorithm is able to not only eliminate the steady state error in the case of conventional PD stand-alone controller but also improve the transient response performance of the system in the case of conventional PID stand-alone controller. Thanks to the SMC theory-based online learning algorithm the parameters of the controller are automatically adapted to cope with the parameter variations and uncertainties. In addition to its robustness, the control approach with the proposed learning algorithm is computationally simple, especially when compared to the gradient descent and the evolutionary algorithms.

\appendices

\section{Proof of Theorem 1}
The time derivative of (\ref{mu}) is written as follows:
\begin{eqnarray}\label{dotmu}
\dot{\mu}_{A_i}(x_1) =-2N_{A_i}(N_{A_i})^{'} \mu_{A_i}(x_1)   \nonumber \\
\dot{\mu}_{B_j}(x_2) =-2N_{B_j}(N_{B_j})^{'} \mu_{B_j}(x_2)
\end{eqnarray}
where
\begin{eqnarray}\label{N}
N_{A_i}=\frac{x_1 - c_{A_i}}{\sigma _{A_i}}   \nonumber \\
N_{B_j}=\frac{x_2 - c_{B_j}}{\sigma _{B_j}}
\end{eqnarray}

The time derivative of (\ref{N}) is written as follows:
\begin{eqnarray}\label{dotN}
\dot{N}_{A_i}=\frac{(\dot{x}_1 - \dot{c}_{A_i})\sigma_{A_i} -(x_1 - c_{A_i})\dot{\sigma}_{A_i}}{\sigma_{A_i} ^{2}}   \nonumber \\
\dot{N}_{B_j}=\frac{(\dot{x}_2 - \dot{c}_{B_j})\sigma_{B_i} -(x_2 - c_{B_j})\dot{\sigma}_{B_j}}{\sigma_{B_j} ^{2}}
\end{eqnarray}

Combination of (\ref{N}) and (\ref{dotN}) gives:
\begin{equation}\label{NdotN}
N_{A_i}\dot{N}_{A_i}=N_{B_j}\dot{N}_{B_j}=\alpha \textrm{sgn}\left(\tau _{c} \right)
\end{equation}

The time derivative of (\ref{layerO3}) is written as follows:
\begin{equation}\label{dotw}
      \dot{\overline{w}}_{ij} = - \overline{w}_{ij} \dot{k}_{ij} +\overline{w}_{ij} \sum\limits_{i=1}^{I}\sum\limits_{j=1}^{J}\overline{w}_{ij} \dot{k}_{ij}
\end{equation}
where
\begin{equation}\label{dotk}
\dot{k}_{ij}=2 \Big(N_{A_i} (N_{A_i})^{'} + N_{B_j} (N_{B_j})^{'} \Big)
\end{equation}

The stability of the proposed control approach is investigated by using the following Lyapunov function:
\begin{equation}\label{V}
V=\frac{1}{2} \left(\tau _{c} \right)^{2}
\end{equation}

The time derivative of (\ref{V}) is written as follows:

\begin{equation}\label{dotV}
\dot{V}=\dot{\tau} _{c} \tau _{c} =\tau _{c} (\dot{\tau} _{n} + \dot{\tau} )
\end{equation}
where
\begin{equation}\label{dottaun}
\dot{\tau} _{n}= \sum\limits_{i=1}^{I}\sum\limits_{j=1}^{J} (\dot{f}_{ij} \overline{w}_{ij} + f_{ij} \dot{\overline{w}}_{ij})
\end{equation}

Substitution of (\ref{dottaun}), (\ref{dotw}), (\ref{dotk}), (\ref{NdotN}) and (\ref{eq5}) into (\ref{dotV}), gives:

\begin{eqnarray}\label{dotV2}
\dot{V} & = & \tau _{c} \Big[\sum\limits_{i=1}^{I}\sum\limits_{j=1}^{J} \Big(\dot{f}_{ij} \overline{w}_{ij} + f_{ij} (- \overline{w}_{ij} \dot{k}_{ij} \nonumber \\
&+&\overline{w}_{ij} \sum\limits_{i=1}^{I}\sum\limits_{j=1}^{J}\overline{w}_{ij} \dot{k}_{ij} ) \Big) + \dot{\tau} \Big] \nonumber \\
  & = & \tau _{c} \Bigg[\sum\limits_{i=1}^{I}\sum\limits_{j=1}^{J} \Bigg(\dot{f}_{ij} \overline{w}_{ij} \nonumber \\
& + & f_{ij} \Big(- 2 \overline{w}_{ij} (N_{A_i} (N_{A_i})^{'} + N_{B_j} (N_{B_j})^{'} ) \nonumber \\
& + & 2\overline{w}_{ij} \sum\limits_{i=1}^{I}\sum\limits_{j=1}^{J}\overline{w}_{ij} (N_{A_i} (N_{A_i})^{'} + N_{B_j} (N_{B_j})^{'} )  \Big) \Bigg) + \dot{\tau} \Bigg] \nonumber \\
  & = & \tau _{c} \Bigg[\sum\limits_{i=1}^{I}\sum\limits_{j=1}^{J} \Bigg(\dot{f}_{ij} \overline{w}_{ij} + f_{ij} \Big(- 4 \overline{w}_{ij} \alpha \textrm{sgn}\left(\tau _{c} \right) \nonumber \\
& + & 4\overline{w}_{ij} \sum\limits_{i=1}^{I}\sum\limits_{j=1}^{J}\overline{w}_{ij} \alpha \textrm{sgn}\left(\tau _{c} \right)  \Big) \Bigg) + \dot{\tau} \Bigg] \nonumber \\
  & = & \tau _{c} \Big[\sum\limits_{i=1}^{I}\sum\limits_{j=1}^{J} \dot{f}_{ij} \overline{w}_{ij}  + \dot{\tau} \Big] \nonumber \\
  & = & \tau _{c} \Big[-\alpha \textrm{sgn}\left(\tau _{c} \right)  + \dot{\tau} \Big] \nonumber \\
  & = & \Big[-\alpha \mid \tau _{c}  \mid  + \dot{\tau}  \mid \tau _{c} \mid \Big]
\end{eqnarray}

The time derivative $\dot{V}$ of the Lyapunov function $V$ must be smaller than zero to satisfy the stability of the learning.
\begin{equation}\label{V3}
\dot{V}= \Big[-\alpha \mid \tau _{c}  \mid  + \dot{\tau}  \mid \tau _{c} \mid \Big] <	0
\end{equation}

When $\dot{\tau}$ reaches its maximal value $B_{\dot{\tau}}$, (\ref{V3}) can be re-written as follows:
\begin{equation}\label{V4}
\dot{V}= \Big[-\alpha \mid \tau _{c}  \mid  + B_{\dot{\tau}} \mid \tau _{c} \mid \Big] < 0 \;\;\;\; if \;\;\;\; B_{\dot{\tau}} < \alpha
\end{equation}

\section{Proof of Theorem 2}

The time derivative of the Lyapunov function in (\ref{lyapunov}) is written as follows:

\begin{eqnarray}\label{}
\dot{V}_{p} & = & \dot{S}_{p}S_{p}=\frac{1}{k^{2}_{d}} \dot{S}_{c}S_{c} = \frac{1}{k^{2}_{d}} \dot{\tau}_{c} \tau _{c}\nonumber \\
& = &  \frac{1}{k^{2}_{d}} \dot{V} < 0, \; \; \forall S_{c}, S_p \neq 0
\end{eqnarray}

\bibliography{ieee}
\bibliographystyle{IEEEtran}

\begin{IEEEbiography}[{\includegraphics[width=1in,height=1.25in,clip,keepaspectratio]{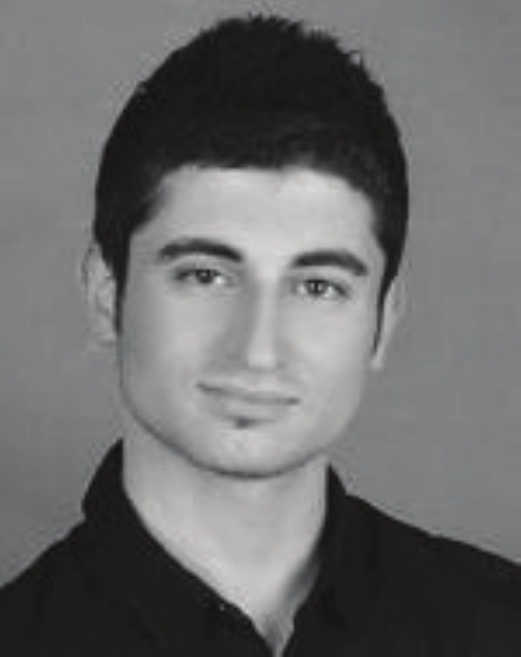}}]{Erkan Kayacan} (S\textquoteright 12) was born in Istanbul, Turkey, on April 17, 1985. He received the B.Sc. and the M.Sc. degrees
in mechanical engineering from Istanbul Technical University, Istanbul, in 2008 and 2010, respectively. He is currently working toward the Ph.D. degree in the Department of Biosystems, KU Leuven.

He is currently a Research Assistant  in KU Leuven at the Department of Biosystems (BIOSYST) in the Division of Mechatronics, Biostatistics and Sensors (MeBioS). His research interests include robotics, nonlinear control, intelligent control, system identification, fuzzy theory, and grey system theory.
\end{IEEEbiography}

\begin{IEEEbiography}[{\includegraphics[width=1in,height=1.25in,clip,keepaspectratio]{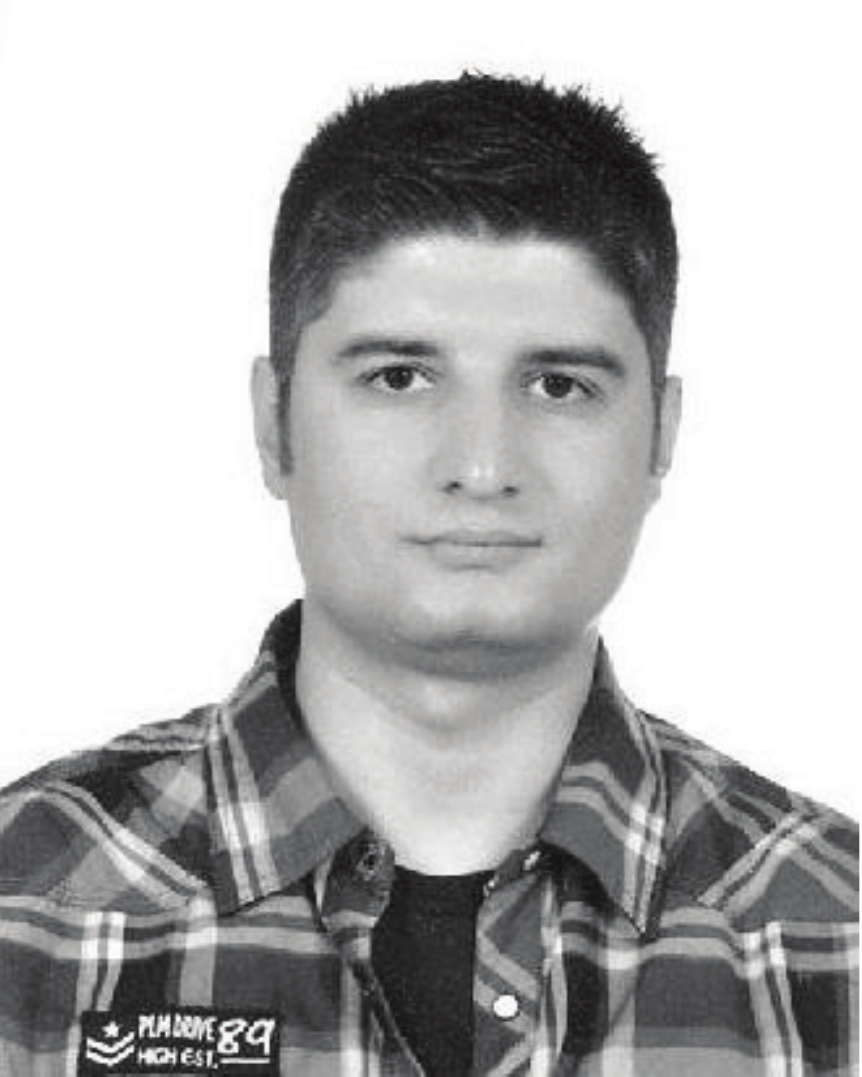}}]{Erdal Kayacan} (S\textquoteright 06-SM\textquoteright 12)  was born in Istanbul, Turkey on January 7, 1980. He received the B.Sc. degree in electrical engineering from Istanbul Technical University, Istanbul, Turkey, in 2003. He received the M.Sc. and Ph.D. degrees in systems and control engineering and electrical and electronics engineering from Bogazici University, Istanbul, Turkey, in 2006 and 2011, respectively. He is currently a post-doctoral researcher in KU Leuven at the department of biosystems (BIOSYST) in the division of mechatronics, biostatistics and sensors (MeBioS). His research interests include large scale systems, soft computing, intelligent control, fuzzy logic theory.

Dr. Kayacan is active in IEEE CIS Student Activities Subcommittee, IEEE CIS Social Media Subcommittee and IEEE SMC Technical Committee on Grey Systems.
\end{IEEEbiography}

\begin{IEEEbiography}[{\includegraphics[width=1in,height=1.25in,clip,keepaspectratio]{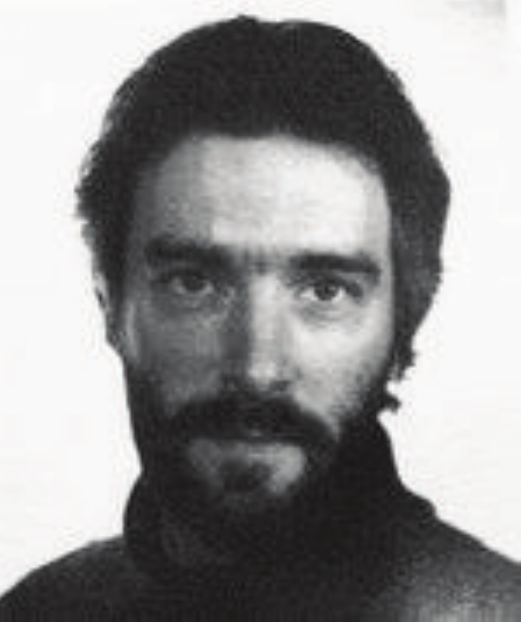}}]{Herman Ramon} graduated as an agricultural engineer from Gent University. In 1993 he obtained a Ph.D. in applied biological sciences at the Katholieke Universiteit Leuven. He is currently Professor at the Faculty of Agricultural and Applied Biological Sciences of the Katholieke Universiteit Leuven, lecturing on agricultural machinery and mechatronic systems for agricultural machinery. He has a strong research interest in precision technologies and advanced mechatronic systems for processes involved in the production chain of food and nonfood materials, from the field to the end user.He is author or co-author of more than 40 papers.
\end{IEEEbiography}

\begin{IEEEbiography}[{\includegraphics[width=1in,height=1.25in,clip,keepaspectratio]{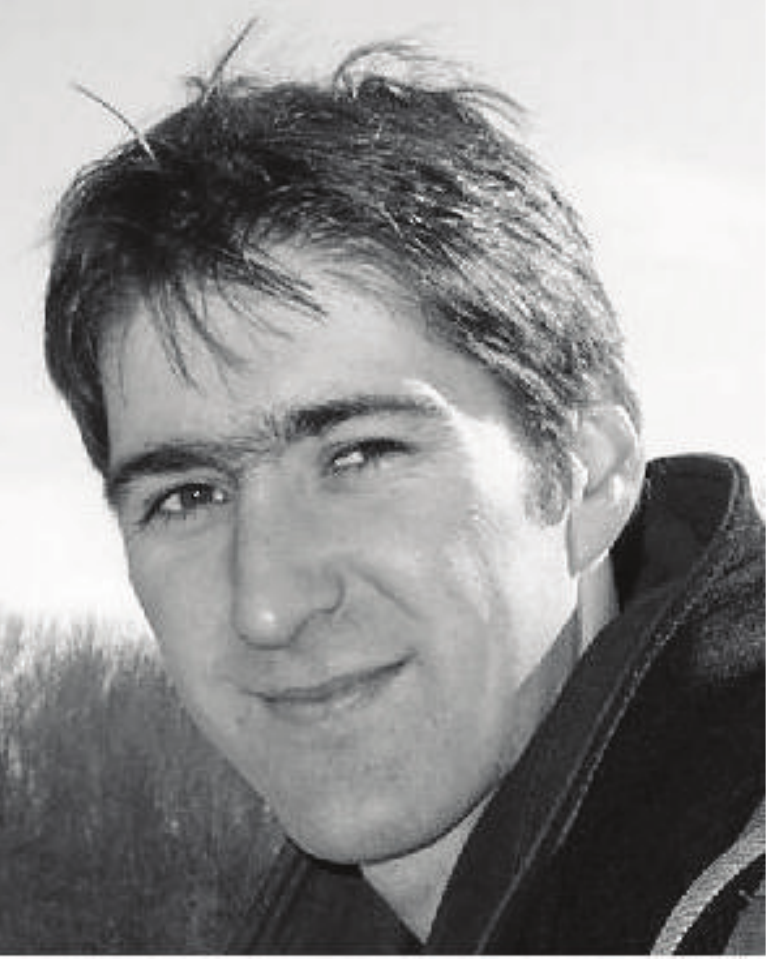}}]{Wouter Saeys} is currently Assistant Professor in Biosystems Engineering at the Department of Biosystems of the University of Leuven in Belgium. He obtained his Ph.D. at the same institute and was a visiting postdoc at the School for Chemical Engineering and Advanced Materials of the University of Newcastle upon Tyne, UK and at the Norwegian Food Research Institute - Nofima Mat in Norway. His main research interests are optical sensing, process monitoring and control with applications in food and agriculture. He is author of 50 articles (ISI) and member of the editorial board of Biosystems Engineering.
\end{IEEEbiography}

\end{document}